\documentclass[letterpaper, 10 pt, conference]{ieeeconf}  

\IEEEoverridecommandlockouts                              

\usepackage{hyperref}
\usepackage{graphicx}
\usepackage{amsmath}
\usepackage{cite}
\usepackage{algorithm}
\usepackage{algpseudocode}
\usepackage{multirow}
\usepackage{gensymb}

\DeclareMathOperator{\arctantwo}{arctan2}

\title{\LARGE \bf{SculptBot: Pre-Trained Models for 3D Deformable Object Manipulation}}

\author{Alison Bartsch$^{1}$, Charlotte Avra$^{1}$, and Amir Barati Farimani$^{1}$
\thanks{$^{1}$With the Department of Mechanical Engineering,
        Carnegie Mellon University \tt\small \{abartsch, cavra, afariman\} @andrew.cmu.edu}}

\begin{document}

\maketitle
\thispagestyle{empty}
\pagestyle{empty}






\begin{abstract}

Deformable object manipulation presents a unique set of challenges in robotic manipulation by exhibiting high degrees of freedom and severe self-occlusion. State representation for materials that exhibit plastic behavior, like modeling clay or bread dough, is also difficult because they permanently deform under stress and are constantly changing shape. In this work, we investigate each of these challenges using the task of robotic sculpting with a parallel gripper. We propose a system that uses point clouds as the state representation and leverages pre-trained point cloud reconstruction Transformer to learn a latent dynamics model to predict material deformations given a grasp action. We design a novel action sampling algorithm that reasons about geometrical differences between point clouds to further improve the efficiency of model-based planners. All data and experiments are conducted entirely in the real world. Our experiments show the proposed system is able to successfully capture the dynamics of clay, and is able to create a variety of simple shapes. Videos and additional figures are available
on our project page at:
\url{https://sites.google.com/andrew.cmu.edu/sculptbot}

\end{abstract}

\section{Introduction}
    
    Deformable objects are prevalent in tasks related to cooking \cite{long2013robotic, dikshit2023robochop, petit2017tracking, mu2019robotic}, manufacturing \cite{kimble2022performance, cortsen2012simulating, lv2022dynamic}, medical procedures \cite{han2020vision, wang2022neural, scheikl2022sim, liu2021real, li2020super}, and more. It is therefore necessary to develop robotic systems that can effectively handle and manipulate these objects before being deployed in these environments. However, due to their mechanical properties, deformable objects permanently deform with robot interaction, increasing dynamic complexity and the difficulty of state estimation. Therefore, unlike rigid object manipulation, common assumptions to estimate the pose and shape of an object from a single viewpoint, even with self-occlusion, cannot be made when perceiving deformable objects, such as clay, that have no inherent shape.
    
    There have been numerous works focused on modeling and manipulating deformable one-dimensional objects (DOOs) or deformable
    linear objects (DLOs) such as rope, cable, sutures, and wires, \cite{saha2007, yu2023coarse, zhu2020, yan2020, zhanng2021, lu2022} and 2D deformable objects such as cloth \cite{huang2023self, yan2021learning, longhini2023edo, jangir20202, borras2020, sunil23a, ha22a}, however systems handling 3D deformable objects, such as modeling clay, remain relatively underexplored, with \cite{shi2022robocraft}, and the follow-up work \cite{shi2023robocook} being the only known soft body manipulation work conducted primarily on physical robot hardware. In this work, we focus on the challenge of predicting how 3D deformable objects, in this case modeling clay, will deform given a robotic action through the task of sculpting clay with a parallel gripper. We propose a novel method leveraging an existing large pre-trained model for point cloud reconstruction, Point-BERT \cite{yu2022point}, to provide a quality latent embedding of the clay states without requiring any training on our particular dataset. We use this embedding to train a latent forward dynamics model to predict the next state of the clay given the current state and a grasp action. The key contributions of this work are as follows:

    \begin{figure}
      \centering
     \includegraphics[width=0.95\linewidth]{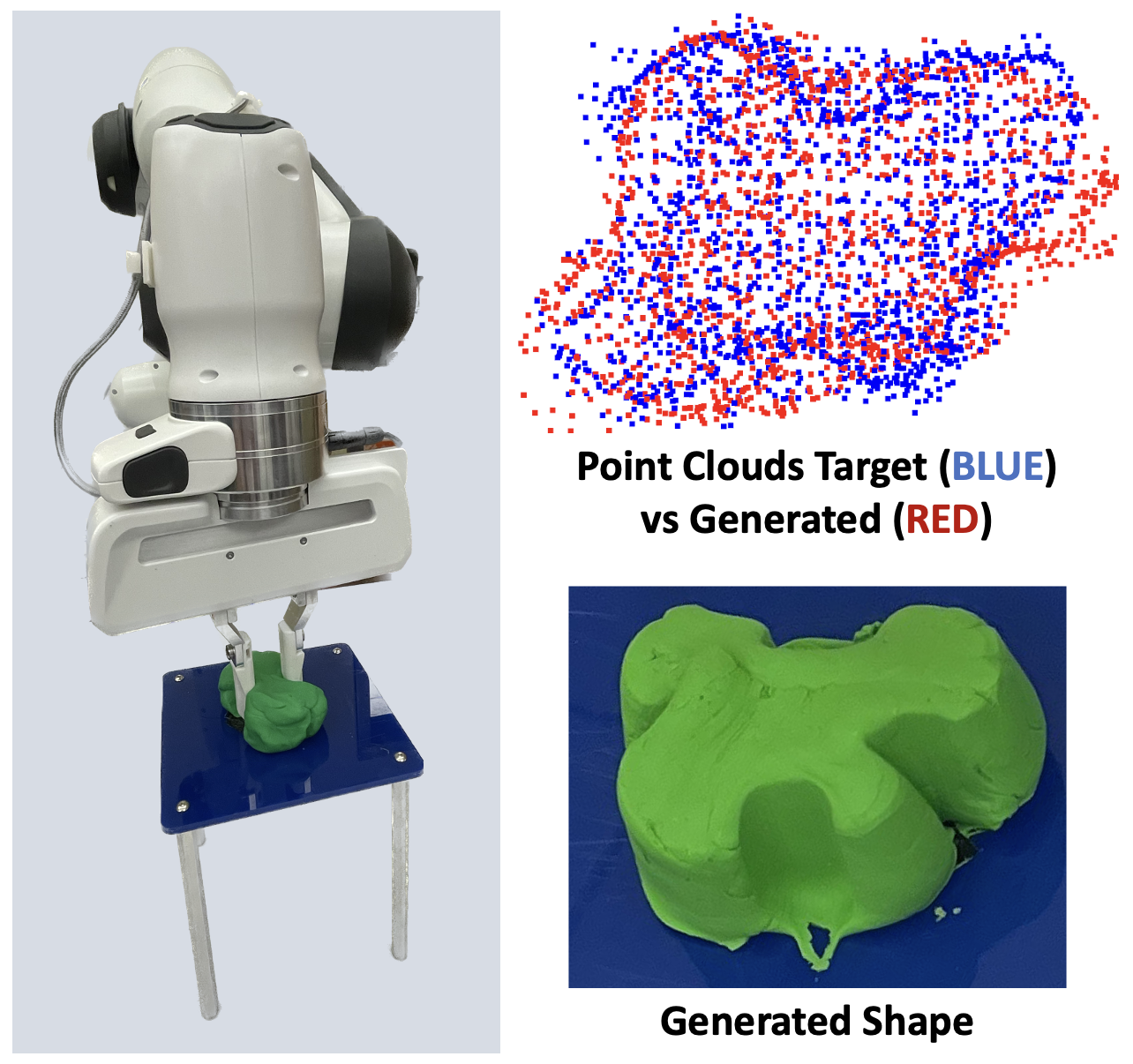}
      \caption{\label{fig:main} SculptBot is a pipeline that leverages the pre-trained point cloud model Point-BERT to learn a latent dynamics model that predicts the next state point cloud of the clay conditioned on the current point cloud and grasp action. When combined with planning methods, SculptBot is able to recreate a variety of 3D shapes. Left shows the robot actively deforming the clay. Right shows the shape the system created compared to the target point cloud.}
      \label{figurelabel}
\end{figure}

    \begin{itemize}
        \item We present the first method to our knowledge leveraging large pre-trained models for deformable object manipulation.
        \item We propose a novel action sampling algorithm that uses the geometrical relationships between point clouds to select candidate actions to achieve the desired deformations.
        \item We present a robust system for sculpting clay with a parallel gripper that is able to successfully replicate target shapes, including those that require changes in thickness. 
    \end{itemize}

\section{Related works}

\textbf{Representations of 3D Data: } Point clouds \cite{rusu20113d} are often used to represent 3D data as they can represent relatively complex objects with a finite number of elements. 3D data can also be represented as voxels \cite{liu2019point}, where a 3D shape is discretized into a set of cubes of a specific size. Voxel grids can then be easily processed using 3D volumetric convolutional neural networks. However, 3D voxel representations tend to be memory intensive and the voxelization process can lead to information loss. A more recent alternative representation are learned signed distance functions (SDF) \cite{park2019deepsdf} which represent the shape with a continuous volumetric field. Researchers have also developed Neural Radiance Fields (NeRFs), neural network-based representations that are able to synthesize novel views of 3D scenes \cite{mildenhall2021nerf}, and follow-up works have extended NeRFs to handle dynamics scenes \cite{pumarola2021d}.

\textbf{Point Cloud Networks: } Point clouds are permutation invariant, and thus special neural networks needed to be designed to handle these inputs. PointNet \cite{qi2017pointnet} and PointNet++ \cite{qi2017pointnet++} use a shared MLP to learn local features of the point cloud. Point transformer \cite{zhao2021point} leverages self attention to learn the local features of the point cloud. More recently, numerous works have created novel self supervised learning methodologies to learn robust encodings of point clouds. PointMAE \cite{pang2022masked} extends the concept of image masking for pretraining into the point cloud domain, where. The follow-up work of PointM2AE \cite{zhang2022point} includes hierarchical masking, motivated by encouraging the model to learn a variety of higher-level and fine-grained features of the point cloud. Point-BERT \cite{yu2022point} devices a BERT-style pretraining method to predict the tokens of the masked patches of the point cloud.

\textbf{Learned Dynamics Models: } There have been many successful works focused on learning dynamics models from data. In \cite{li2019learning}, researchers learn particle-based dynamics models for a variety of materials as an alternative to particle-based simulators. Similarly, in \cite{li2022graph}, researchers learn a graph-based particle dynamics model for fluids. Learned dynamics models are also prevalent within the domain of model-based reinforcement learning. In DREAMER \cite{hafner2019dream}, and the follow-up DREAMER-V2 \cite{hafner2020mastering}, a learned dynamics model is used to predict the agent's future success by leveraging dynamics model rollouts.

\textbf{Manipulation of Deformable Objects: } The task of manipulating deformable objects has been investigated previously. In RoboCraft \cite{shi2022robocraft}, researchers are similarly learning a dynamics model to deform clay into target shapes. However, in this work the authors focus their evaluation on a 2D alphabet test set. We hope to explore a wider variety of more complex, 3D target shapes in this work. In an extension to RoboCraft, researchers present RoboCook \cite{shi2023robocook}, which uses a similar dynamics model with a set of diverse tools to create the 2D alphabet shapes. In \cite{yan2021learning}, researchers train a latent dynamics model with contrastive learning to handle 2D deformable objects, such as cloth. A similar task of folding cloth was explored in \cite{li2015folding}. In \cite{wi2022virdo}, researchers focus on handling semi-rigid deformable objects, such as silicone spatulas. Beyond the tasks of handling deformable objects, research has also focused on building better simulators, such as SoftGym \cite{lin2021softgym}, or PlasticineLab \cite{huang2021plasticinelab}, learning material properties such as elasticity \cite{frank2010}, and improving the possibility of sim-to-real transfer with these complex objects \cite{matas2018sim}.

\section{Methodology}

Our method primarily consists of three modules - the vision system, the pre-trained tokenizer from Point-BERT \cite{yu2022point}, and the learned latent dynamics model. One of the key considerations with this work was the choice of representation for the 3D shape of the clay. Point clouds were a natural choice due to the richness of information, while still minimizing memory usage. The choice of state representation for the clay directly informed both the vision system as well as the dynamics model. Point clouds are an unstructured representation, which requires a special neural network architecture that can handle this. 

\begin{figure*}
      \centering
     \includegraphics[width=1.0\linewidth]{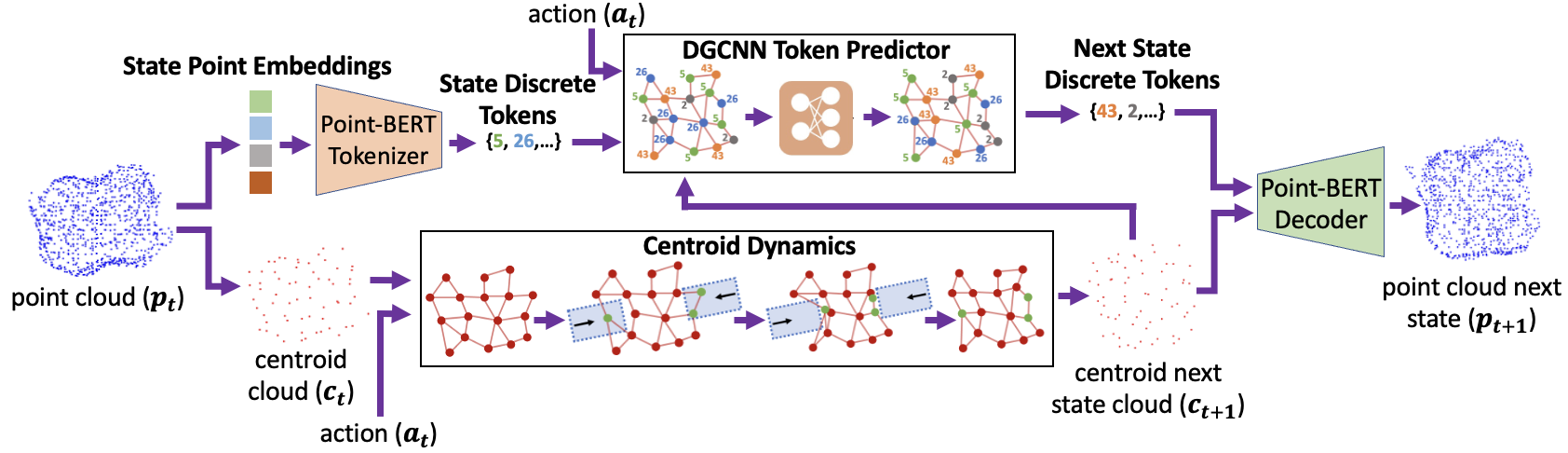}
      \caption{\label{fig:pipeline} The entire dynamics prediction pipeline. We first use farthest point sampling and k-nearest neighbors to cluster the original point cloud into 64 clusters. These 64 clusters become a much smaller and sparser point cloud. We then use the pre-trained dVAE from Point-BERT to tokenize each cluster. The centroid point cloud is passed through a simple physics-based dynamics approximator to predict the next state centroid point cloud given the grasp action.  This predicted next state centroid point cloud is passed to the point token predictor dynamics model along with the state centroid tokenization and the grasp action. The point token predictor predicts the tokens for each next state centroid, which represent the geometrical structure of the points within that region of the cloud. These predicted tokens along with the predicted centroid point cloud are then passed through the dVAE decoder to reconstruct the full dense predicted next state point cloud. }
      \label{figurelabel}
  \end{figure*}

\subsection{Vision System}

    Our vision system consists of 4 Intel RealSense D415 RGB-D cameras (shown in Figure \ref{fig:hardware}). We found that the standard technique for calculating extrinsics between cameras using a calibration checkerboard pattern resulted in relatively inaccurate combined point clouds. Instead, we used an asymmetrical 3D object to calculate the extrinsics between cameras using global point cloud registration, specifically RANSAC \cite{fischler1981random}, and fine-tuned them with the Iterative Closest Point algorithm (ICP) \cite{besl1992method}. This significantly improved our final fused point cloud accuracy, to a fusal with approximately 0.5 cm error comparable to that of RoboCraft's vision module \cite{shi2022robocraft}. 

\begin{figure*}
      \centering
     \includegraphics[width=0.98\linewidth]{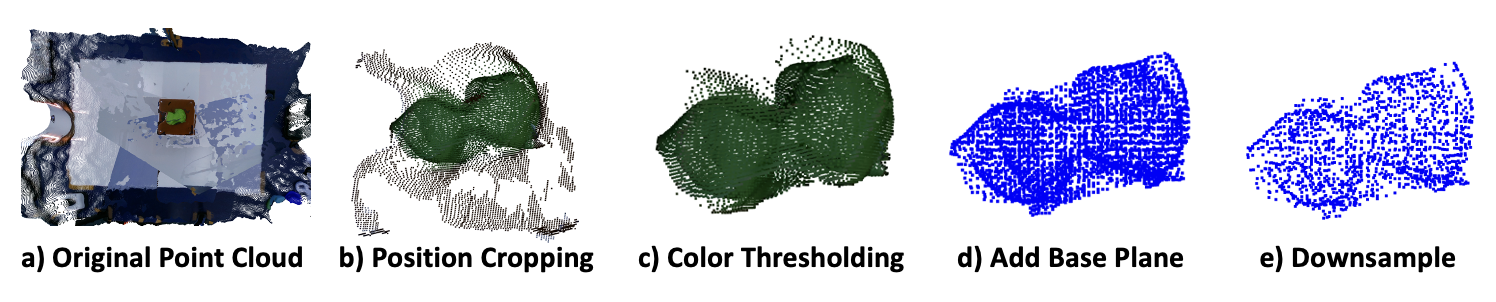}
      \caption{\label{fig:pcl_processing} The full preprocessing pipeline for the point clouds. a) The original point cloud of the scene. b) the scene after position-based cropping to eliminate the elevated stage. c) The point cloud after color-based thresholding. d) The point cloud after removing statistical outliers and adding in a base plane. e) The point cloud downsampled to 2048 points.}
      \label{figurelabel}
\end{figure*}

    Once we have a quality 3D point cloud of the environment, we perform a simple position-based crop to isolate the elevated stage (shown in Figure \ref{fig:hardware}), and use color-based cropping in the LAB colorspace to isolate the clay from the table. We then extract the points closest to the base of the clay (by indexing points below a z threshold). We use this base shape outline to crop a plane of points to form the bottom of the clay. We combine this base plane with the original cloud to form a fully enclosed point cloud shell. This processing is based on the assumption that the clay is always resting on the elevated stage, thus the base of the clay will be at that z-position. Once we have the clay shell, we downsample the point cloud to 2048 points. A full visualization of the point cloud preprocessing pipeline is shown in Figure \ref{fig:pcl_processing}.

\subsection{Pre-Trained Point-BERT Tokenizer}

\begin{figure}
      \centering
     \includegraphics[width=0.93\linewidth]{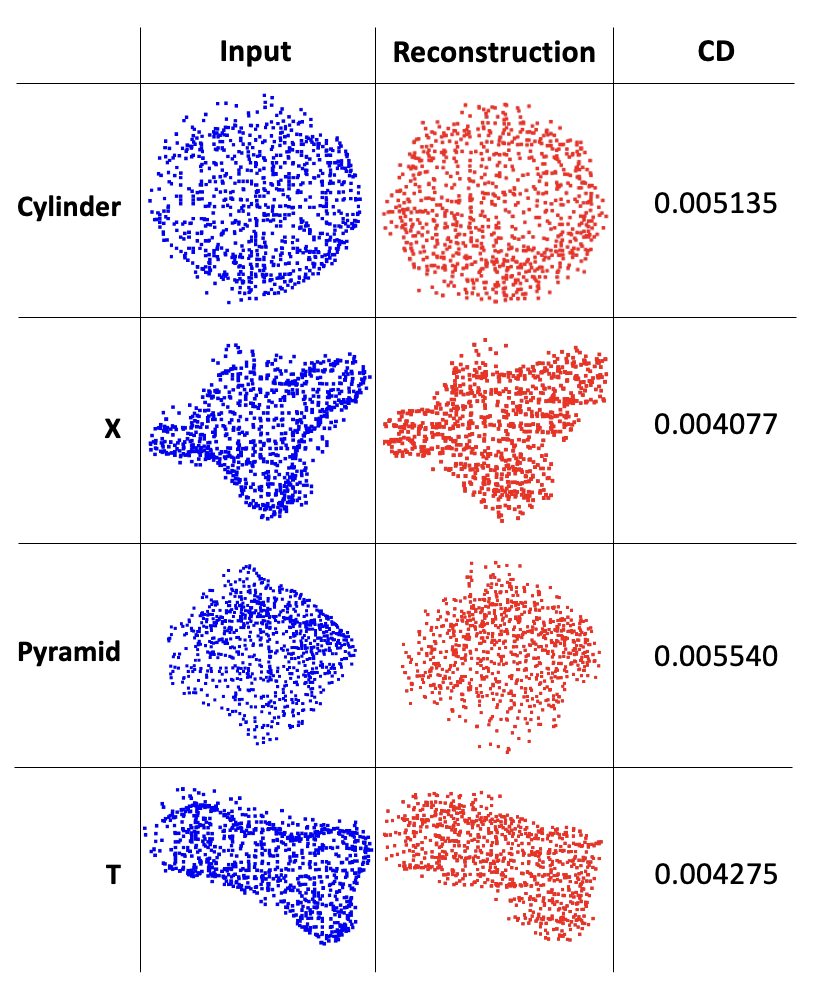}
      \caption{\label{fig:dvae_recon} The dVAE from Point-BERT provides quality reconstruction of the real-world clay point clouds without requiring any finetuning on our dataset. This allows us to train a latent dynamics model predicting the material deformation in the Point-BERT embedding space. }
      \label{figurelabel}
\end{figure}

One of the challenges with manipulating 3D deformable objects is that models handling 3D representations typically require a significant amount of data to properly train. However, when working within the robotic space, collecting sufficient data can be incredibly time consuming. To address this challenge, we leverage the tokenizer from Point-BERT, which is pre-trained on the ShapeNet dataset \cite{chang2015shapenet}, a collection of 3D point clouds of 3,135 common categories. We find in practice that the discrete variational autoencoder (dVAE) from Point-BERT learned a sufficient latent representation, that needs no fine-tuning on the clay-specific dataset. A visualization of the reconstruction quality of our real-world clay data is shown in Figure 
\ref{fig:dvae_recon}.

Point-BERT is originally a pre-training strategy for point cloud transformers. First, the point clouds are grouped into local patches, or clusters using farthest-point sampling and and k-nearest neighbors. Each cluster has a positional embedding to maintain the regional locations for reconstruction. The points in each local patch are subtracted with the centroid position to eliminate bias and only represent the regional point cloud structure. Each cluster is then passed through a small PointNet \cite{qi2017pointnet} to output a discrete point embedding that describes the general point cloud shape within that region. These discrete tokens and the positional embeddings are then passed through the decoder to reconstruct the original point cloud. In this work, we use the dVAE tokenizer from Point-BERT and train a dynamics model in the latent space of the positional and point embeddings.

\subsection{Latent Dynamics Model}

Given the latent representation from the dVAE of Point-BERT consists of two components (positional embedding of the clusters, and the point tokens for each cluster), we developed two seperate dynamics modules to propagate the dynamics of a grasp action. First, we developed a simple physics approximator that propagates the point cloud based on the trajectory of the gripper fingertips during the grasp. Next, based on the next state centroids and the grasp action, we train a DGCNN \cite{phan2018dgcnn} token predictor model to predict how each cluster's point token changes given the grasp action. When combined with the pre-trained dVAE to encode the state point cloud and reconstruct the predicted cloud, the system predicts the next state of the clay point cloud given the current state point cloud and the grasp action. The grasp action is defined as the x,y,z position of the end-effector, the rotation about the z-axis, and the distance between the fingertips.

\subsubsection{Centroid Dynamics}

Each cluster is passed through a physics-based dynamics approximator that propagates the point clouds based on the grasp action. This dynamics approximator is very light weight, moving the points that are inliers in the approximated gripper trajectory mesh based on the grasp pose. These points are moved normal to the surface of the gripper the distance between their initial location and the gripper final position. Additionally, we impose some distance constraints to the point cloud, ensuring that the points cannot exceed a pre-specified distance. After we approximate the point motion due to the grasp, we update the point cloud to reinforce these distance constraints. While this dynamics approximator is relatively naive, when paired with the second stage learned dynamics model we expect it to account for the simplicity.

\subsubsection{DGCNN Point Token Dynamics}

The token dynamics model takes the predicted next state positional embeddings from the centroid dynamics model, the state point token embeddings and the grasp action as input to predict the next state point tokens that minimizes the Chamfer distance between the reconstructed predicted next state and ground truth point clouds. The Chamfer distance is the sum of squared distances between the nearest neighbor correspondences between point clouds, and is a common point cloud reconstruction metric. The predicted next state point tokens are passed back through the dVAE decoder to reconstruct the full predicted next state point cloud. 

\begin{equation}
    d_{CD}(X,Y) = \sum_{x \in X} \min_{y \in Y} ||x - y||_2 ^2 + \sum_{y \in Y} \min_{x \in X} ||x - y||_2 ^2  
\end{equation}

\begin{algorithm}
\caption{\label{alg:logic}Geometric Sampler}
\label{figurelabel}
\begin{algorithmic}
\State \textbf{Input:} $P_{state}$ \Comment{current state point cloud}
\State \textbf{Input:} $P_{target}$ \Comment{target shape point cloud}
\State \textbf{Input:}$N_{clusters}$ \Comment{algorithm parameter}
\State \textbf{Input:} $EE_{width}$ \Comment{hardware parameter}

\State $\mu_{state} = KMeans(P_{state}, N_{clusters})$ 
\State $\mu_{target} = KMeans(P_{target}, N_{clusters})$ 

\State $\Vec{\delta} = (\mu_{target} - \mu_{state})$

\State $dists = |\mu_{state, i} - \mu_{target, i}| $


\State Sample pair of centroids with probability $p = \frac{dists}{\sum dists}$
\State $(x,y,z) = \mu_{state} + \frac{1}{2} \Vec{\delta} EE_{width}$
\State $r_z = \arctantwo (\Vec{\delta}_y, \Vec{\delta}_x)$
\State $d = \frac{1}{2}|\mu_{target} - (x,y,z)|$

\State $actions = [x, y, z, r_z, d]$

\State \textbf{Return:} $actions$

\end{algorithmic}
\end{algorithm}

\subsection{Geometry-Informed Action Sampling}

We developed an action sampler that leverages geometric knowledge of the point clouds to efficiently sample quality potential actions. The concept of the geometric sampler would be to identify the regions of the current point cloud that are most different from the target cloud, and search for actions that would apply a change to the state in those areas. Especially as the action space increases in dimensionality, it is important to have a more efficient action sampling method than random parameter selection. The full algorithmic description of the action sampler is shown in Algorithm \ref{alg:logic}. We first utilize k-means \cite{macqueen1967some} to cluster each point cloud into $N$ regions. We then pair the clusters between the two point clouds based on those with the smallest Euclidean distance. The cluster pairs are sampled with a probability proportional to the distance, and an action is generated to push the cluster in the state point cloud towards that in the target point cloud.

\section{Experimental Setup Details}

\begin{figure}
      \centering
     \includegraphics[width=0.93\linewidth]{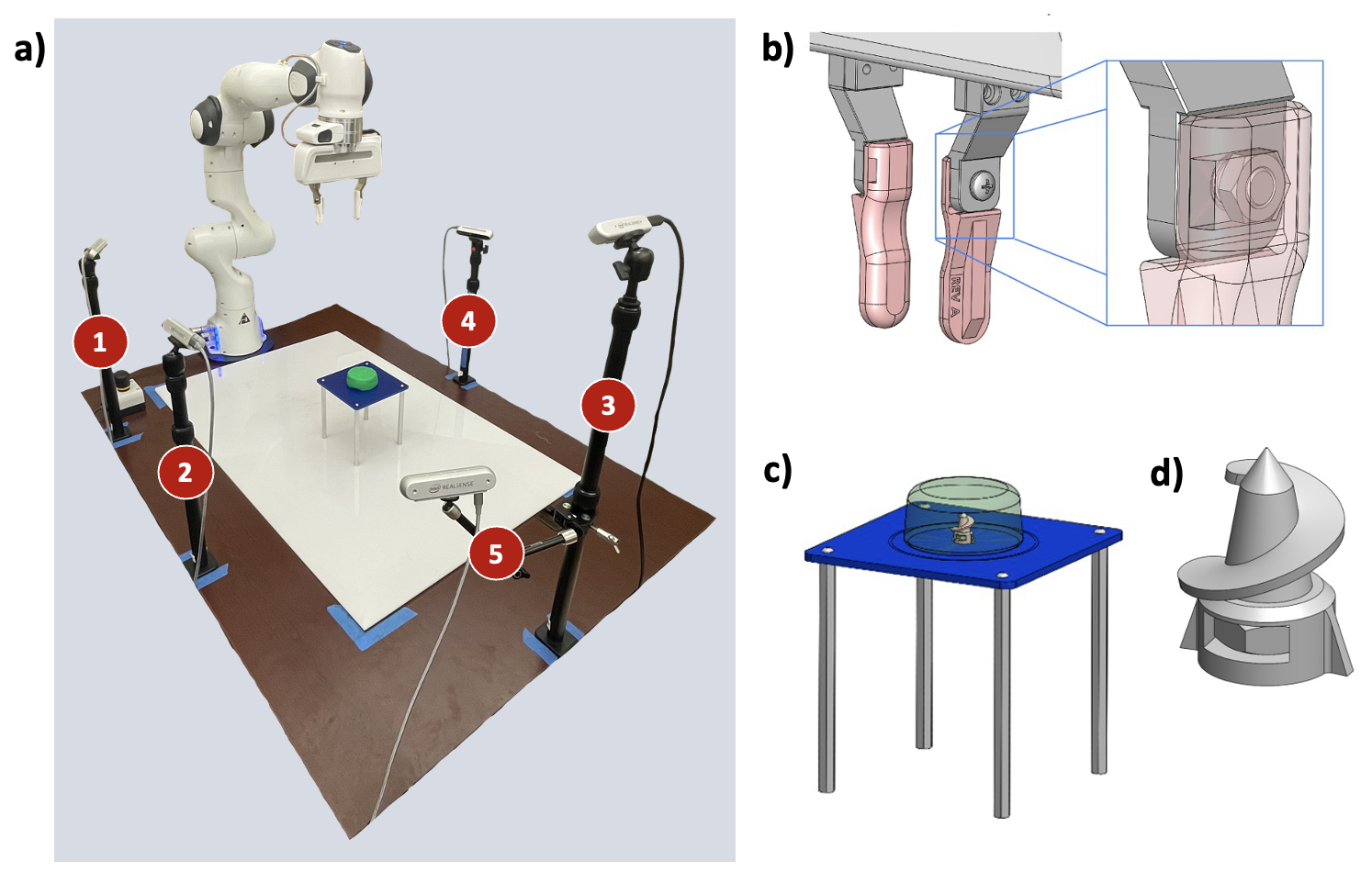}
      \caption{\label{fig:hardware}The hardware setup: a) general workspace with 4 Intel RealSense D415 cameras for 3D scene reconstruction (1-4), and one Intel RealSense D455 camera for capturing video (5), b) the custom 3D printed fingertips, c) the elevated stage, d) screw-in anchor for the clay on the elevated stage.}
      \label{figurelabel}
\end{figure}

All data collection and experiments were conducted on a physical 7-axis Franka Emika Panda manipulator equipped with a two-finger parallel gripper. There are 4 Intel RealSense D415 cameras mounted to the workspace. The clay sits atop an elevated stage in the center of the workspace, and is loosely attached to the stage with a small screw to ensure the clay generally remains on the stage. Details of the hardware setup are shown in Figure \ref{fig:hardware}. We assume the clay volume remains consistent throughout data collection, always initialized to a cylinder 6 cm in diameter and 2.5 cm in height. 

\subsection{Data Collection}

We collect two separate datasets to train alternate versions of the latent dynamics model, 1) a random action dataset, 2) a human demonstration dataset. The random action dataset is collected by randomly sampling action parameters and executing the generated grasps. The point cloud of the clay is collected before and after each grasp. We consider a grasp trajectory to be 10 consecutive grasps applied to the clay before the clay is re-initialized to the starting shape. We collect a total of 15 trajectories, consisting of approximately 30 minutes of data collection time. To increase the size of our dataset, we augment the dataset by applying varying rotation transforms about the z-axis to the point cloud (in intervals of 6$\degree$) and the grasp action. This augmentation strategy is sound, as we have the assumption that the clay always remains fixed on the elevated stage surface. The human demonstration dataset is collected with kinesthetic teaching where the human controls the end-effector position, rotation and the distance between the fingertips. We chose this over alternative human demonstration collection techniques, such as \cite{george2023openvr}, because we found kinesthetic teaching produced more accurate sculptures. The human operator collected 5 trajectories of varying length for creating a cone, cylinder, line, pyramid, T, and X. The same rotation augmentation technique is applied to these trajectories. While this is much more time consuming, we collected this dataset to explore differences in model performance trained on each dataset.

\section{Results}

Once we have the trained latent dynamics model, we can plan action trajectories to reach a variety of target point clouds. In section \ref{sec:dynamics_eval} we first evaluate the quality of the learned latent dynamics model. In section \ref{sec:teleoperation} we have a human subject teleoperate the robot to generate a variety of shape targets to investigate the difficulty of the task with the robot embodiment. Finally, in section \ref{sec:shape_targets} we deploy the learned dynamics model with model predictive control for the sculpting task. We particularly evaluate the performance of the dynamics model trained on the random and human demonstration datasets, as well as the proposed geometric sampling strategy compared to random shooting.

\subsection{Evaluation of the Dynamics Model}
\label{sec:dynamics_eval}

\begin{figure}
      \centering
     \includegraphics[width=0.93\linewidth]{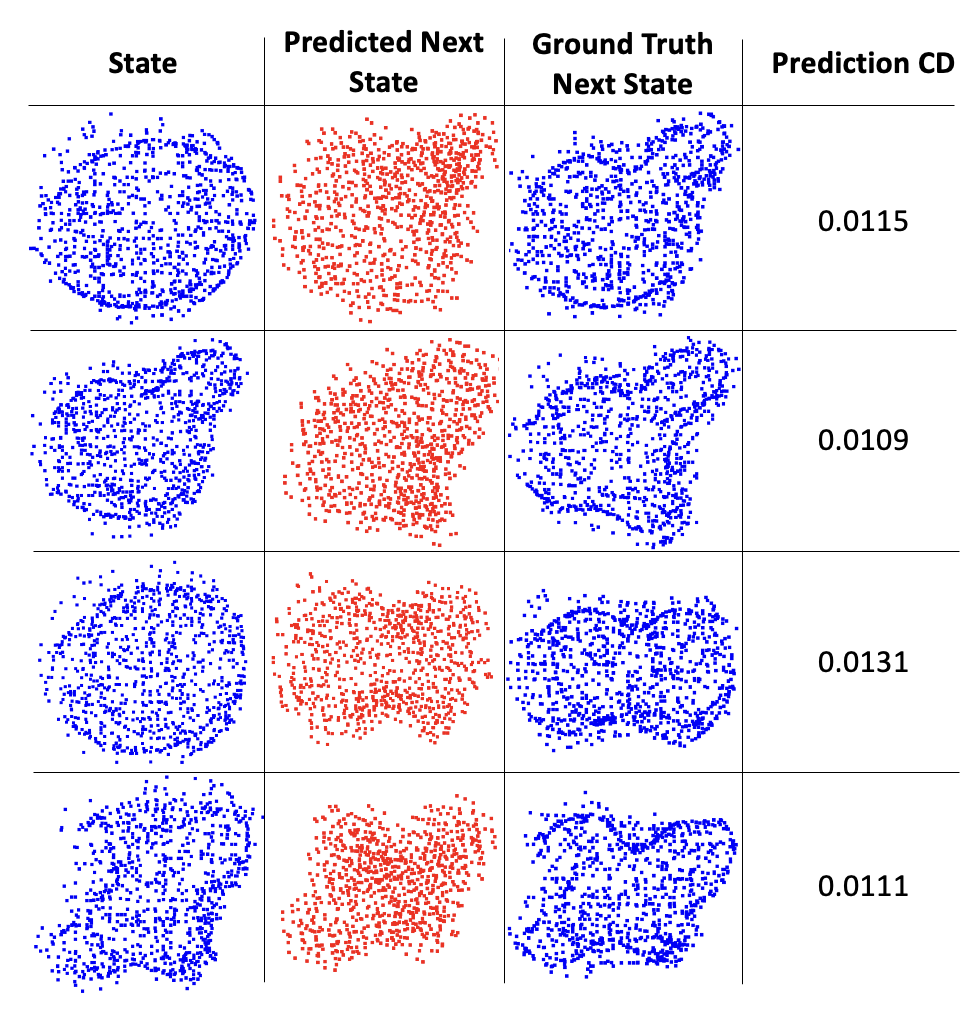}
      \caption{\label{fig:dynamics_pred} A visualization of some next state predictions on the test set by the latent dynamics model trained on the human demonstration dataset. It is clear the model is able to capture and predict the large geometric changes caused by various grasp actions. However, some of the details may not be captured, likely due to the shape reconstruction, as the quality appears similar to some of the detail lost during reconstruction, shown in Fig \ref{fig:dvae_recon}. This loss is a side effect of leveraging the pre-trained model from Point-BERT, and is not sufficient to justify training our own point cloud encoder, as it would require substantially more data.}
      \label{figurelabel}
\end{figure}

To evaluate the quality of the next state point cloud predictions of our dynamics model, we report the Chamfer Distance (CD) between the predicted next state and ground truth next state of the combined test sets of the human demonstration and random action datasets. A visualization of the predictions for a few states are shown in Figure \ref{fig:dynamics_pred}. The performance of the dynamics models on the entirety of the dataset are shown below in Table \ref{tab:cds}. The dynamics model trained on the human demonstration dataset had a lower mean Chamfer distance and lower variance for the next state predictions. However, this difference is relatively small, and may not be great enough to motivate the additional human effort of collecting the demonstrations. 

\begin{table}[]
\caption{Chamfer Distance (CD) for the dynamics models trained on the different datasets collected.} \label{tab:cds}
\centering
\begin{tabular}{@{\extracolsep{\fill}}lll}
        \hline
         \textbf{ } & \textbf{Centroid Next State CD} & \textbf{Full Next State CD} \\ 
        \hline
        \hline
        \multirow{1}{*}{\textbf{Human Demos}} & 0.0450 $\pm$ 0.0069 & 0.0109 $\pm$ 0.0024 \\
        \hline
        \multirow{1}{*}{\textbf{Random Actions}} & 0.0451 $\pm$ 0.0069 & 0.0113 $\pm$ 0.0028 \\
        \hline 
    \end{tabular}
\end{table}

\subsection{Human Teleoperation Baseline}
\label{sec:teleoperation}

    The task of deforming a small block of clay with a parallel gripper is incredibly difficult, as a relatively simple action results is complicated changes to the clay shape. To better understand what quality of results we could expect for this task, we conducted human teleoperation experiments. We used a simple kinesthetic teaching system where the human controls the end-effector position, rotation, and the distance between the fingertips. The results of this experiment are shown in Figure \ref{fig:human_baseline} and are compared to the results of our dynamics system. Although sculpting with a parallel gripper can be unintuitive, the human teleoperators were able to successfully create a range of simple shapes.

\subsection{Hardware Deployment}
\label{sec:shape_targets}

\begin{figure}
      \centering
     \includegraphics[width=0.77\linewidth]{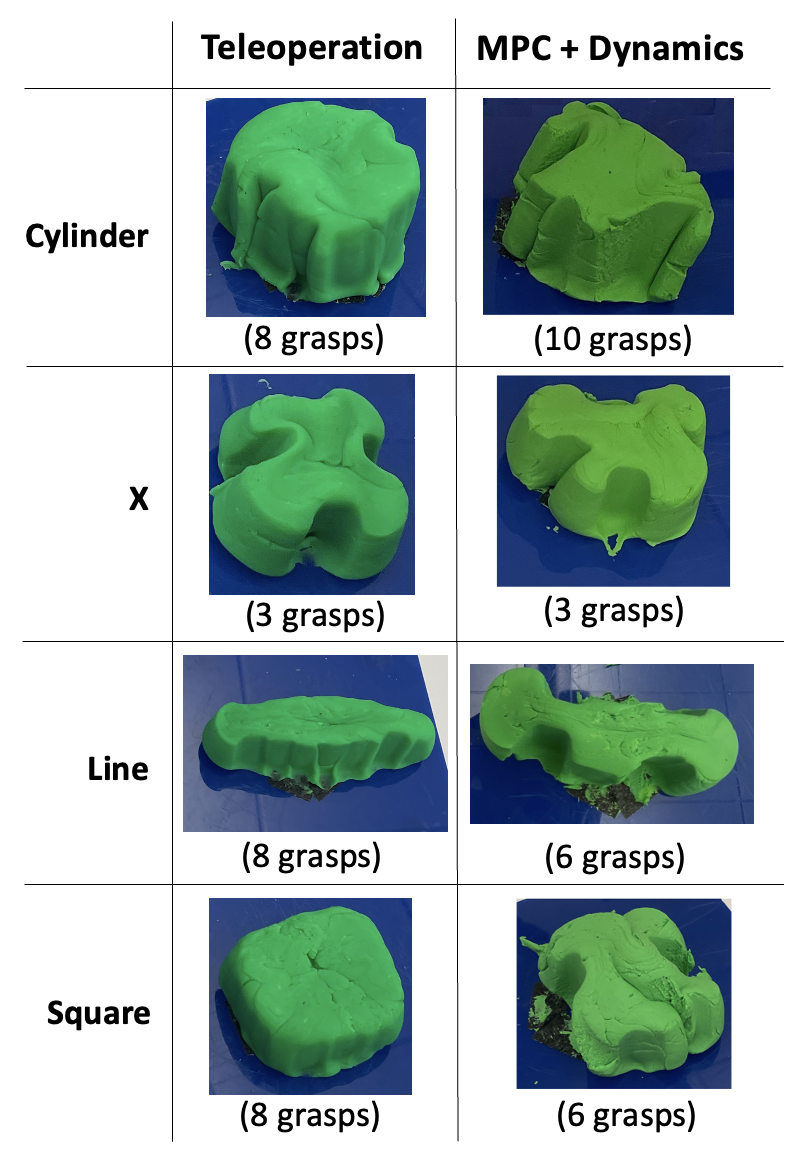}
      \caption{\label{fig:human_baseline}The shapes a human was able to create when guiding the robot and its parallel gripper (Left) compared to the shapes the human demonstration trained dynamics model combined with MPC and geometric sampling was able to create (Right). Visually, it is clear that our system is not able to perform better than human oracle, but based on the reconstruction metrics, it is able to successfully recreate the key structural aspects of the shapes.}
      \label{figurelabel}
  \end{figure}

We deploy the fully trained dynamics model on hardware for the sculpting task by unrolling trajectories with model predictive control (MPC). For these experiments, we are planning one step into the future, and selecting the action that minimizes the Chamfer distance between the expected next state and the target point cloud. We first compare the performance of the proposed geometric action sampler as compared to random shooting (shown in Table \ref{tab:sampler}). For these experiments, we are using the dynamics model trained on the human demonstration dataset. For the random shooting action sampling, we randomly sample 2500 actions for MPC, whereas with the geometric-based sampler we only need to sample 35 actions. For each comparison, we run the system 3 times for each target, and report the mean and standard deviation. The geometric-based sampler produces sculptures of similar quality compared to random shooting, while being significantly more sample efficient. Additionally, the geometric-based sampler achieves these results with on average fewer grasp actions as compared to random shooting. Next, we evaluate the performance of the models trained on the human demonstration and random datasets with MPC with geometric-based sampling (shown in Table \ref{tab:datasets}) on a variety of target shapes. The model trained with the human demonstration dataset performs substantially better on a few of our test shapes, particularly 'X' and 'triangle'. Our human demonstration dataset did include examples of humans creating an 'X', but not 'triangle'. As the human demonstration dataset was used to train a single-step dynamics model, this means that the model trained on this data is able to more accurately predict the dynamics of the actions that are in fact useful for creating these shapes. In the future, it would be interesting to explore a middle-ground dataset collection technique that is more intelligent than random actions, but not as time consuming as requiring human demonstrators. A visualization of the quality of some of the shapes the system is able to create are shown in Figure \ref{fig:human_baseline}. Our system is able to reliably generate replications of target sculptures with reasonable Chamfer distance that capture the key structural aspects of the target shape.

\begin{table}[]
\caption{Comparing sculpting performance of the proposed geometric action sampling as compared to random shooting.}\label{tab:sampler} 
\centering
\begin{tabular}{@{\extracolsep{\fill}}lllll}
        \hline
         \textbf{Target} & \textbf{Sampler} & \textbf{\# Grasps} & \textbf{CD} \\ 
        \hline
        \hline
        \multirow{2}{*}{\textbf{X}} & Random & 4.2 $\pm$ 0.84 & 0.0024 $\pm$ 0.0003 \\
         & Geometric & 3.6 $\pm$ 0.89 & 0.0029 $\pm$ 0.0007\\
        \hline
        \multirow{2}{*}{\textbf{T}} & Random & 10.7 $\pm$ 3.1 &  0.0069 $\pm$ 0.0012 \\
         & Geometric & 6.7 $\pm$ 3.5 & 0.0058 $\pm$ 0.002 \\
        \hline 
        \multirow{2}{*}{\textbf{Square}} & Random & 3.7 $\pm$ 0.6  & 0.0044 $\pm$ 0.0005 \\
         & Geometric & 4.7 $\pm$ 1.5 & 0.0042 $\pm$ 0.0004 \\
        \hline 
        \multirow{2}{*}{\textbf{Line}} & Random & 10.7 $\pm$ 3.1 & 0.00739 $\pm$ 0.001 \\
         & Geometric & 5.2 $\pm$ 2.5 & 0.0068 $\pm$ 0.0015 \\
         \hline
         \multirow{2}{*}{\textbf{Cylinder}} & Random &  11.5 $\pm$ 2.1 & 0.0027 $\pm$ 0.0003 \\
         & Geometric & 12.7 $\pm$ 2.5 & 0.0041 $\pm$ 0.0002 \\
         \hline
         \multirow{2}{*}{\textbf{Triangle}} & Random & 4.0 $\pm$ 2.0 & 0.0048 $\pm$ 0.0009 \\
         & Geometric & 2.3 $\pm$ 0.6 & 0.0031 $\pm$ 0.0005 \\
         \hline
    \end{tabular}
\end{table}

\begin{table}[]
\caption{Comparing sculpting performance of the dynamics models trained on the random action dataset and the human demonstration dataset with the geometric sampler.}\label{tab:datasets} 
\centering
\begin{tabular}{@{\extracolsep{\fill}}llll}
        \hline
         \textbf{Target} & \textbf{Dataset} & \textbf{\# Grasps} & \textbf{CD} \\ 
        \hline
        \hline
        \multirow{2}{*}{\textbf{X}} & Human Demo & 
        3.0 $\pm$ 0.0  & 0.0025 $\pm$ 0.0007 \\
         & Random & 3.7 $\pm$ 0.6 & 0.0041 $\pm$ 0.0005 \\
        \hline
        \multirow{2}{*}{\textbf{T}} & Human Demo & 6.7 $\pm$ 3.5 & 0.0058 $\pm$ 0.0018 \\
         & Random & 4.0 $\pm$ 2.6 & 0.0060 $\pm$ 0.0013 \\
        \hline 
        \multirow{2}{*}{\textbf{Square}} & Human Demo & 4.7 $\pm$ 1.5 & 0.0042 $\pm$ 0.0004 \\
         & Random & 3.3 $\pm$ 0.6 & 0.0039 $\pm$ 0.0011 \\
        \hline 
        \multirow{2}{*}{\textbf{Line}} & Human Demo & 5.3 $\pm$ 2.5 & 0.0068 $\pm$ 0.0015 \\
         & Random & 3.3 $\pm$ 0.6 & 0.0054 $\pm$ 0.0007 \\
         \hline
         \multirow{2}{*}{\textbf{Cylinder}} & Human Demo & 12.7 $\pm$ 2.5 & 0.0041 $\pm$ 0.0002 \\
         & Random & 10.5 $\pm$ 0.7 & 0.0051 $\pm$ 0.0015 \\
         \hline
         \multirow{2}{*}{\textbf{Triangle}} & Human Demo & 2.3 $\pm$ 0.6 & 0.0031 $\pm$ 0.0005 \\
         & Random & 3.3 $\pm$ 1.2 & 0.0071 $\pm$ 0.0015 \\
         \hline
    \end{tabular}
\end{table}

\section{Conclusion}

    In this work we present SculptBot, a system that leverages pre-trained point cloud reconstruction models to learn a latent dynamics model for 3D deformable object manipulation. Through our experiments, we demonstrate that our model is able to successfully capture the dynamics of the clay, and is able to create a variety of simple shape sculptures when combined with model predictive control and a geometric-based action sampling algorithm. One of the key limitations of this work is that a parallel gripper may not be the best tool for the sculpting task. From our human teleoperation experiments, the quality of the shapes our human oracle was able to create are relatively coarse and simplistic. In future work, we hope to investigate alternative sculpting actions, first by expanding the types of actions the robot can take, such as pinch and twist, smooth, etc. Additionally, we hope to provide the robot with a set of sculpting tools and incorporate actions for both additive and subtractive clay sculpting.

\newpage
\bibliographystyle{IEEEtran}
\bibliography{ref}

\end{document}